\documentclass{llncs} 

\usepackage{amsmath,amssymb,mathtools} 
\usepackage{multirow}
\usepackage{tabularx,graphicx}
\usepackage{booktabs}
\usepackage{hyperref}
\usepackage{color,soul}

\title{ Efficient moving point handling for incremental 3D manifold reconstruction}
\author{Andrea Romanoni \and Matteo Matteucci}

\institute{DEIB, Politecnico di Milano,
  Via Ponzio 34/5, 20133, Milano, Italy
        {\tt\small \{andrea.romanoni,matteo.matteucci\}@polimi.it}
}

\DeclareMathOperator*{\argmin}{arg\,min}
\begin{document}

\maketitle

\begin{abstract}
As incremental Structure from Motion algorithms become effective, a good sparse point cloud representing the map of the scene becomes available frame-by-frame.
From the 3D Delaunay triangulation of these points, state-of-the-art algorithms build a manifold rough model of the scene.
These algorithms integrate incrementally new points to the 3D reconstruction only if their position estimate does not change.
Indeed, whenever a point moves in a 3D Delaunay triangulation, for instance because its estimation gets refined, a set of tetrahedra have to be removed and replaced with new ones to maintain the Delaunay property; the management of the manifold reconstruction becomes thus complex and it entails a potentially big overhead.
In this paper we investigate different approaches and we propose an efficient policy to deal with moving points in the manifold estimation process. We tested our approach with four sequences of the KITTI dataset and we show the effectiveness of our proposal in comparison with state-of-the-art approaches.
\end{abstract}

\section{Introduction}
\label{sec:intro}
Incremental 3D reconstruction from a sparse point cloud is gaining interest in the computer vision community as incremental Structure from Motion algorithms are consolidating  \cite{wu13}. 
This is clearly true for those applications where a rough, but dense, surface represents a sufficient and effective representation of the scene, e.g, for traversability analysis in unmanned vehicle navigation. 
Furthermore, in real-time applications, the map of the environment needs to be updated online, and the surface has to be estimated incrementally. 

Most of the existing algorithms \cite{Lovi_et_al_11,Pan_et_al09,Litvinov_Lhuillier_13,litvinov_Lhiuller14} bootstrap the reconstruction of a mesh surface from the 3D Delaunay triangulation of a sparse point cloud. Indeed, the 3D Delaunay triangulation  is very powerful:
the Delaunay property, i.e., no point of the triangulation is inside the sphere circumscribing any tetrahedron, avoids as much as possible the resulting tetrahedra to have a degenerate shape \cite{Maur_02}; it is self-adaptive, i.e., the more the points are dense the more the tetrahedra are small; it is very fast to compute, and to  update against point removal or addition; off-the-shelf libraries, such as CGAL \cite{cgal}, enable a very simple and efficient management of it. 

As soon as a Delaunay triangulation is available, several approaches exist to extract a surface taking into account the visibility of each point. 
The simplest algorithm is the Space Carving \cite{Kutulakos_Seitz05}: it initializes all the tetrahedra as \emph{matter}, then it marks as \emph{free space} the tetrahedra intersected by the camera-to-point \emph{viewing rays}, i.e., the lines from the camera center to the observed 3D points in the triangulation. 
The boundary between free space and matter represents the final surface of the scene.
Pan et al. \cite{Pan_et_al09} improve upon this simple procedure by proposing an online probabilistic Space Carving, but this is not an incremental approach: they start from scratch every time new points are added.
Lovi et al. \cite{Lovi_et_al_11} present the first incremental Space Carving algorithm which runs real-time, but, as for the previous methods, the estimated surface is not guaranteed to be manifold 

Several reasons lead to enforce the manifold property as explained in \cite{lhuillier20142}. 
Most Computer Graphics algorithms need the manifold property, for instance smoothing with Laplace-Beltrami operator \cite{Meyer03}, or the linear mesh parametrization \cite{saboret00}.
Moreover the manifold property enables surface evolution in mesh-based Multi-View Stereo, as in \cite{vu_et_al_2012,Delaunoy_et_al_08}.the manifold property enables a photometric refinement by surface evolution such as with the high accurate Multi-View Stereo mesh-based algorithm as in \cite{vu_et_al_2012,Delaunoy_et_al_08}.
With these approaches is hard to estimate the surface evolving flow in the presence of non manifold vertices: indeed they compute for each vertex the gradient minimizing the reprojection error, by summing-up the contribution of the incident facets; if the vertex is not manifold, this gradient does not converge. As a further proof of this, \cite{vu_et_al_2012} needs to manually fix the surface estimated via s-t cut.
As in \cite{vu_et_al_2012}, it is possible to fix the mesh as a post-processing step, but reconstructing directly a manifold as in the proposed paper, enables the design of a fully automatic pipeline which do not need human intervention.

In literature, the only algorithm reconstructing a manifold incrementally was proposed by Litvinov and Lhuiller \cite{Litvinov_Lhuillier_13,litvinov_Lhiuller14}. 
In their work, the authors  bootstrap from the Space Carving procedure and, by taking into account the number of intersections of each tetrahedron with the viewing rays, they reconstruct a surface keeping the manifold property valid. 
The main limitation is that  Litvinov and Lhuiller insert a point into the Delaunay triangulation only when its position is definitive, then they cannot move the point position anymore even in the case they could refine their estimate. 
The main reason of Litvinov and Lhuiller design choice has to be ascribed to the computational cost of updating the visibility information along the viewing rays incident to each moved point, and the computational cost of updating part of the Delaunay triangulation, which in turn induces a new manifold reconstruction iteration step.

\begin{figure}[t]
\centering
\begin{tabular}{cccccc}
\includegraphics[width=0.16\columnwidth]{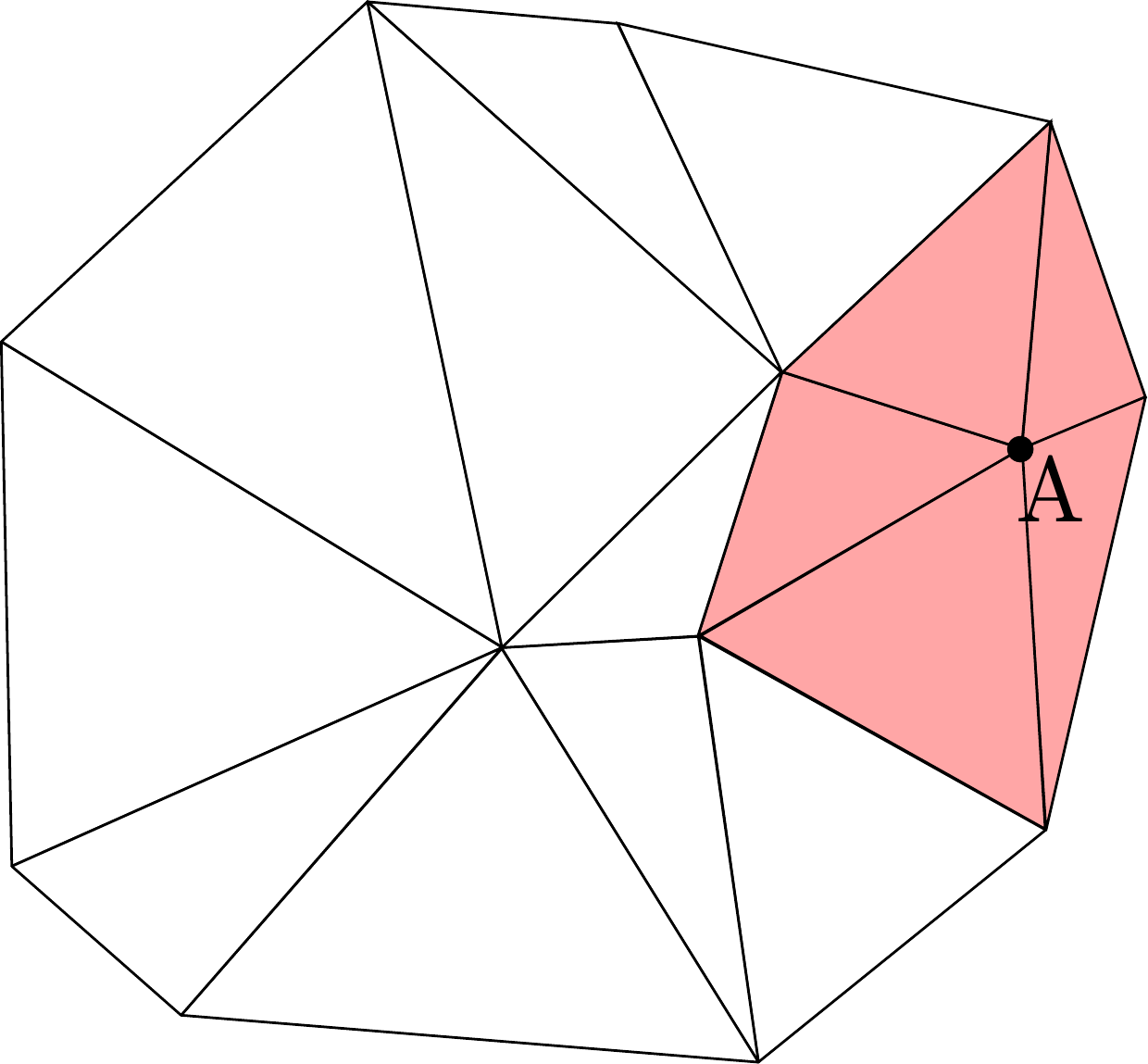}&
\includegraphics[width=0.16\columnwidth]{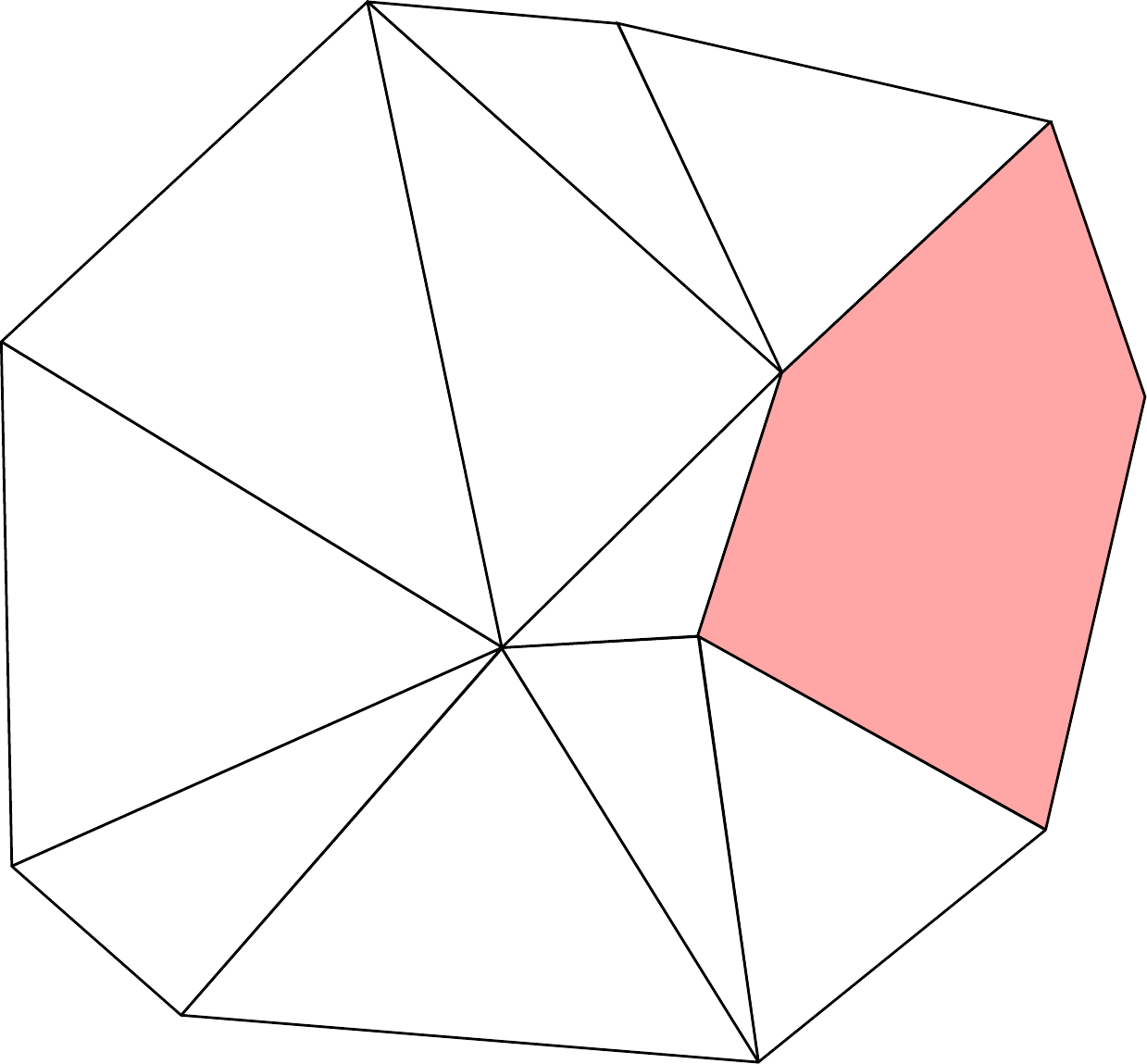}&
\includegraphics[width=0.16\columnwidth]{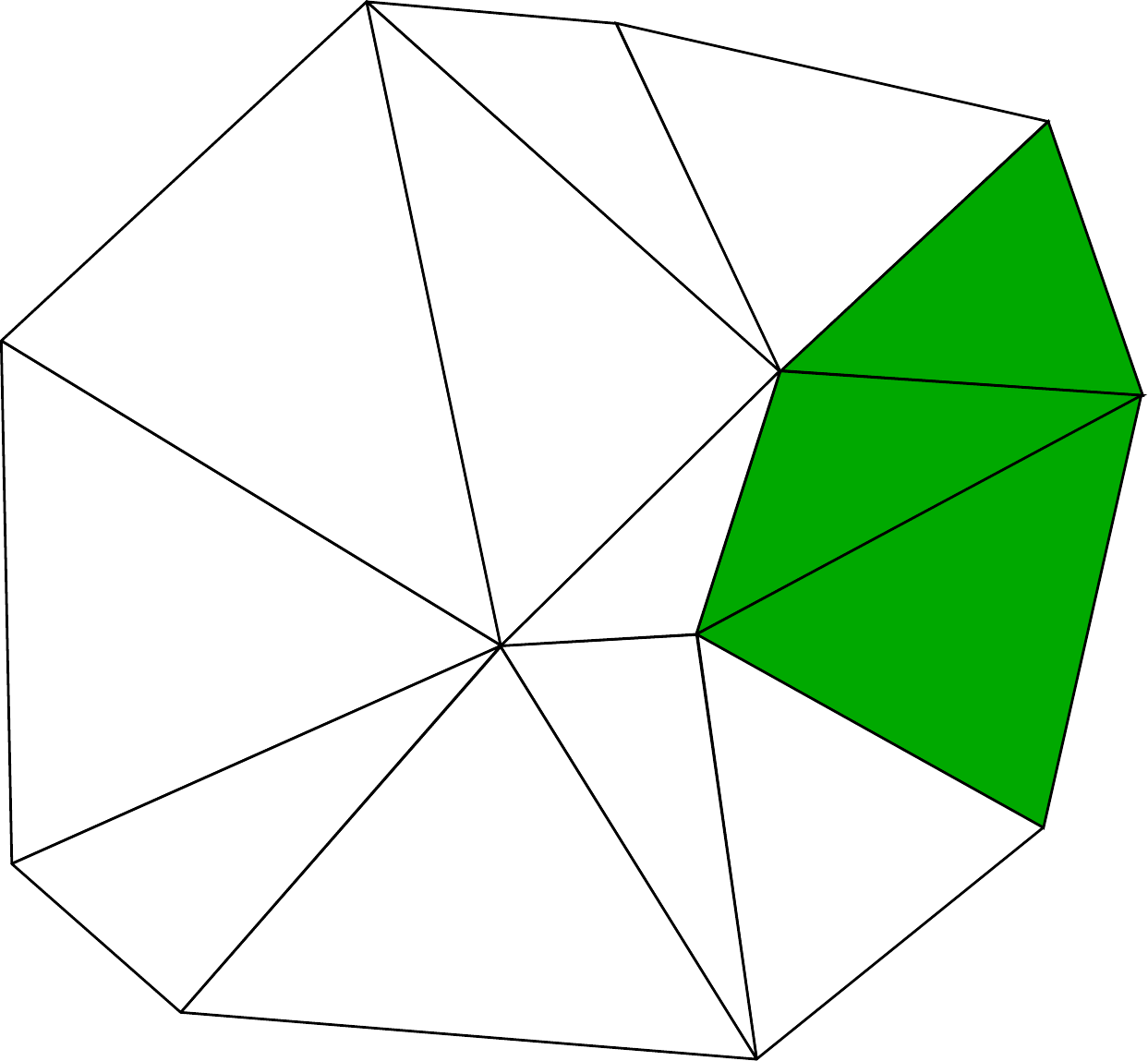}&
\includegraphics[width=0.16\columnwidth]{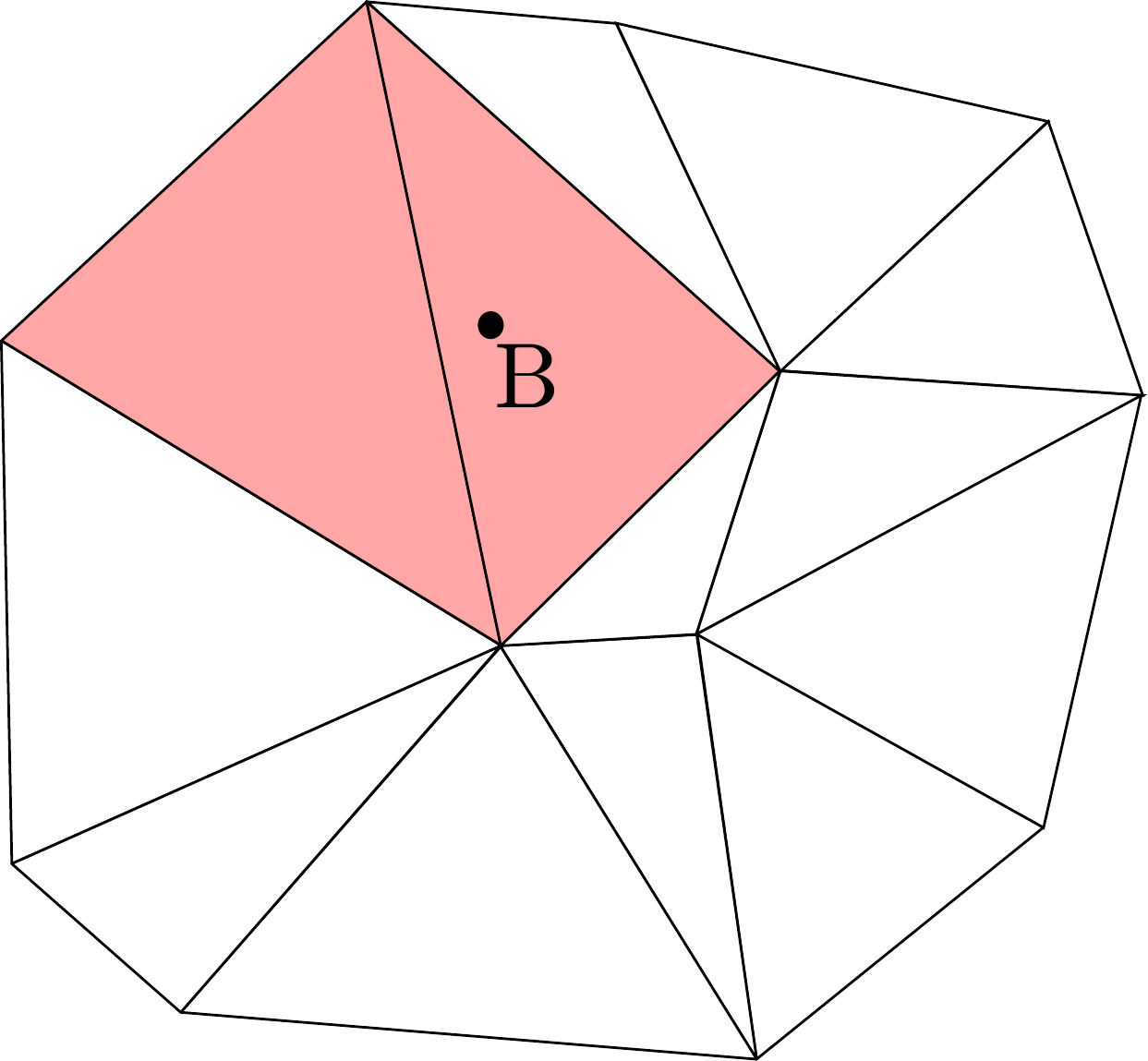}&
\includegraphics[width=0.16\columnwidth]{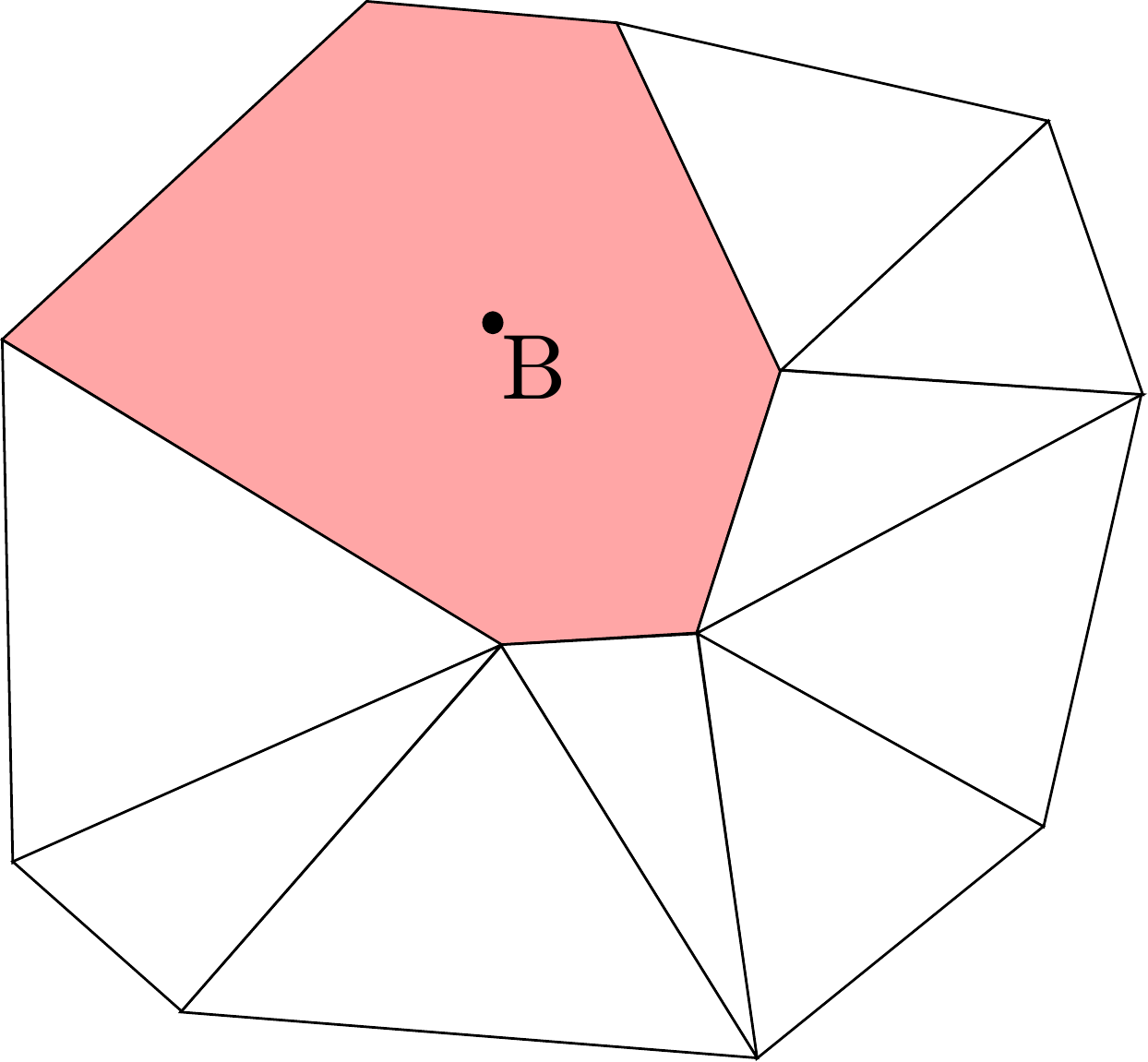}&
\includegraphics[width=0.16\columnwidth]{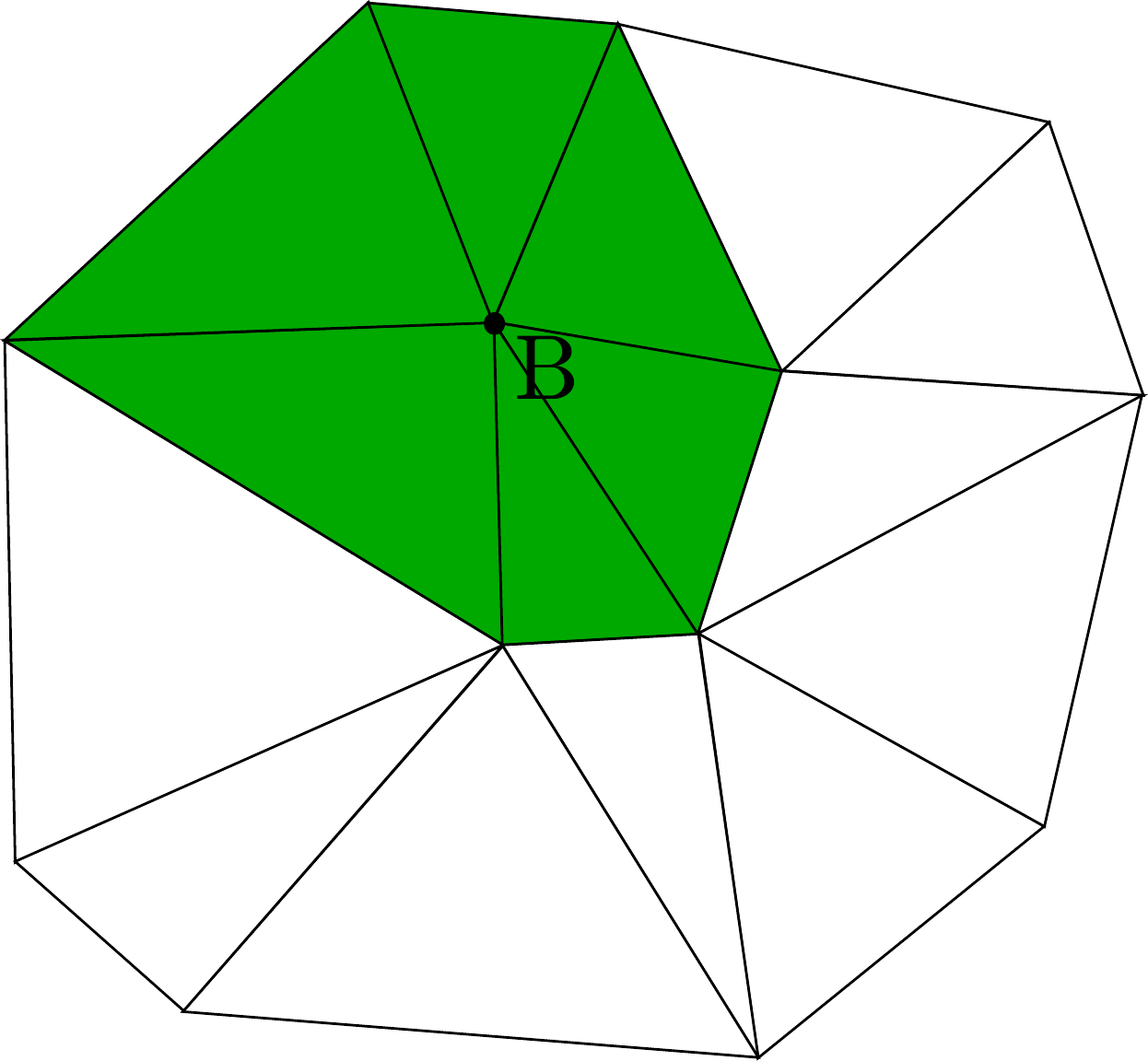}\\
(a)&(b)&(c)&(d)&(e)&(f)
\end{tabular}
\caption{Example of point removal in 2D case. Light red triangles depict are removed and replaced with the new dark green ones.}
\label{fig:moving}
\end{figure}

Indeed, the very common approach to deal with a point moving in the triangulation, is to remove it and add it back in the new position \cite{cgal} (Fig. \ref{fig:moving}). 
When we remove a point (the point A in Fig. \ref{fig:moving}(a)) and we want to keep the Delaunay property, we have to remove all the tetrahedra incident to that point (light red triangles in Fig. \ref{fig:moving}(b)); then, we add a new set of tetrahedra to triangulate the resulting hole (dark green triangles in \ref{fig:moving}(c)).
When we add a new point into the triangulation  (the point B in Fig. \ref{fig:moving}(d)), a set of tetrahedra would conflict with it, i.e., the Delaunay property is broken (light red triangles in Fig. \ref{fig:moving}(d)); so, we remove this set of tetrahedra again (red triangles in Fig. \ref{fig:moving}(e)) and we add a new connected set that re-triangulate the hole (dark green triangles in Fig. \ref{fig:moving}(f)).
Whenever a set of tetrahedra is replaced, we have to transfer conveniently the information about the visibility (matter or free space) of the removed tetrahedra to the new one. 
In addition to this, we have to update the visibility of the tetrahedra crossed by a visibility ray from one camera to the moved point.
For these reasons the update of the point position is computational demanding.

To complete the overview of the incremental reconstruction methods from sparse data, we mention here another very different approach was proposed by Hoppe et al. \cite{Hoppe13} who label the tetrahedra with a random field, and extract the surface via graph-cuts by minimizing a visibility-consistent energy function. This incremental algorithm is effective and handles the moving points, but the manifold property of the reconstructed surface is not yet guaranteed.

In this paper we propose, to the best of our knowledge, the first manifold 3D reconstruction algorithm from sparse data which deals with dynamic point changes. In particular, we  show that in this setting the algorithm by Lovi et al. \cite{Lovi_et_al_11} provides a feasible solution, but it is very inefficient and we propose a novel efficient policy to handle the visibility update of Delaunay tetrahedra with moving points. 

In Section \ref{sec:manifold} we summarize a slightly modified version of the approach of \cite{Litvinov_Lhuillier_13} we use to reconstruct a manifold surface.
In Section \ref{sec:3D-Reconstruction} we describe Lovi's approach \cite{Lovi_et_al_11} and our proposal to deal with moving points, together with a complexity analysis that explains why our approach is more efficient. 
In Section \ref{sec:experimental-results} we show the experimental results on the publicly available dataset KITTI \cite{Geiger_et_al12}, while, in Section \ref{sec:conclusion} we point out some future works and in  the conclusion of the paper.

\section{Manifold Reconstruction} 
\label{sec:manifold}
In this paper we reconstruct a manifold surface that represents the observed scene. 
A surface is manifold if and only if the neighborhood of each point is homeomorphic to a disk. 
In the discrete case, the points are the vertexes of a mesh, and the neighborhood is represented by the incident triangles (or polygons); a surface is manifold if each vertex $v$ is \emph{regular}, i.e., if and only if the edges opposite to $v$ form a closed path without loops  (see \cite{Litvinov_Lhuillier_13} for more details). 

%
%

\subsection{Incremental manifold extraction with tetrahedra weighting}
\label{subsec:incrementalManifold}

In this section we briefly summarize our variation on the method originally proposed in \cite{Litvinov_Lhuillier_13} enhanced by a weighting scheme that avoids the creation of most visual artifact in the final mesh (more discussion about visual artifacts in \cite{litvinov_Lhiuller14}).  
In our Space Carving algorithm, a weight roughly represents how many rays intersect a tetrahedron, and in the following, a tetrahedron belongs to \emph{free space} if its weight $w$ is higher than a threshold $T_w$ (in our case $T_w = 1.0$).

\subsubsection{Sparse point cloud}
The input of our algorithm is a sparse 3D point cloud, estimated incrementally by assuming the camera poses to be known.
For each keyframe, i.e., every $K=5$ frames, we extract Edge-point features, i.e. 2D points laying on the image edges \cite{Rhein_et_al13}; these points represent measurements of 3D points.
Frame-by-frame we track these features with the Kanade-Lucas-Tomasi tracker, and, at each keyframe, we estimate the new positions of the 3D points with the new  measurements available from the tracking. New estimates are obtained by triangulating the 2D tracked points and by minimizing the reprojection error with a Gauss-Newton algorithm. Once a new estimate of a 3D point is available, we add it to the Delaunay triangulation, i.e., to the reconstruction; then we update its position according to the new measurements: this update induces the motion of the points inside the triangulation.

\subsubsection{3D reconstruction}
The reconstruction of the surface bootstraps from the manifold partitioning the 3D triangulation between the set $O$ of \emph{outside} tetrahedra, i.e., the manifold subset of the free space (not all the free space tetrahedra will be part of the manifold), and the complementary set $I$ of inside tetrahedra, i.e. the remaining tetrahedra that represent the matter together with the free space tetrahedra which would invalidate the manifold property.

Let  $\delta (O_{t_{\text{init}}})$ be the initial manifold. This initial manifold is obtained with the following steps. \emph{Point Insertion}: add all the 3D points estimated up to time $t_{\text{init}}$ and build thir 3D Delaunay triangulation. \emph{Ray tracing and tetrahedra weighting}: for each viewing ray, the algorithm traverses the triangulation and adds a weight $w_1 = 1.0$ to the intersected tetrahedra, a weight $w_2 = 0.8$ to the neighbors and a weight $w_2 = 0.2$ to the neighbors of their neighbors. 
Such weighting scheme acts as a smoother of the visibility and avoids the creation of visual artifacts; it is the main difference between our algorithm and the algorithms proposed in \cite{Litvinov_Lhuillier_13,litvinov_Lhiuller14}. \emph{Growing}: initialize a queue $Q$ starting from the tetrahedron with the higher weight. Then: (a) pick  the tetrahedron with highest weight from $Q$ and add it to $O_{t_{\text{init}}}$ only if the resulting surface between $O_{t_{\text{init}}}$ and $I_{t_{\text{init}}}$ remains manifold; (b) if inserted  add the neighboring tetrahedra to the queue $Q$, otherwise discard it; continue iteratively until $Q$ is empty.

Once the system is initialized, a new set of points $P_{t_k}$ is estimated at each $t_k= t_{\text{init}} + k*T_k$ frame, named keyframes, where $k \in \mathbb{N^+}$ and $T_k$ is the inverse of the keyframe rate. 
The insertion of a point $p\in P_{t_k}$ causes the removal of the set $D_{t_k}$ of tetrahedra breaking the Delaunay property, and, the surface $\delta (O_{t_k}) = \delta (O_{t_{k-1}} \setminus D_{t_k})$ is not guaranteed to be manifold anymore. 
To avoid this, the authors in \cite{Litvinov_Lhuillier_13} define a list of tetrahedra $E_{t_k} \supset D_{t_k}$ and apply the \emph{Shrinking} procedure, i.e., the inverse of Growing:  they subtract iteratively from $O_{t_{k-1}}$ the tetrahedra  $\Delta \in E_{t_k}$ keeping the manifoldness valid.
After this process, it is likely that $D_{t_k} \cap O_{t_k} = \emptyset$. Whenever $D_{t_k} \cap O_{t_k} \neq \emptyset$ the point $p$ is not added to the triangulation, i.e., is dropped.
Once all points in $P_{t_k}$ have been added (or dropped), the growing process runs similarly to the initialization procedure, but the queue $Q$ is initialized with the tetrahedra $\Delta \in T \setminus O$ such that  $\Delta \cap \delta O \neq \emptyset$.

\section{Reconstructing a manifold with moving points}
\label{sec:3D-Reconstruction}
As previously described, Litvinov and Lhuiller \cite{Litvinov_Lhuillier_13} algorithm adds points to the triangulation only when their 3D position is completely defined; by doing this, there are no changes in the Delaunay triangulation, induced by moving points. This  results in a restriction if we would like to refine the estimation of the position of a point 3D position after its insertion.

Only Lovi et al. \cite{Lovi_et_al_11} presents an incremental Space Carving algorithm which deals with moving points, but their method does not enforce the manifold property.
In this paper we verify the approach of Lovi et al. \cite{Lovi_et_al_11} to be very inefficient for manifold reconstruction, and we present a different approach to deal with moving points that leads to a significant faster computation.

\subsection{The straightforward approach}
\label{subsec:straightforward_way}
The simplest way to deal with moving points while reconstructing a manifold surface, is to apply a straightforward modification to the so called \emph{Refinement Event Handler} by Lovi et al. in \cite{Lovi_et_al_11}.
The Refinement Event Handler algorithm assumes that, for each tetrahedron in the Delaunay triangulation a list of the intersecting viewing rays is stored. In our voting schema an intersecting ray is each ray that increase the weight of the tetrahedron. 

Let $p_{\text{old}}$ be a point that moves to position $p_{\text{new}}$, the algorithm in \cite{Lovi_et_al_11} moves the point by removing point $p_{\text{old}}$ and adding $p_{\text{new}}$ as a new point, according to the classical approach of \cite{Devillers03}, then for each point they apply the following steps. \emph{Rays collection}: collect in a set $U$ all the rays stored into the tetrahedra incident to $p_{\text{old}}$, i.e., those affected by the $p_{\text{old}}$ removal (e.g., the light red triangle in Fig. \ref{fig:moving}(a)). \emph{Vertex removal}: remove the vertex $p_{\text{old}}$ and its neighboring tetrahedra from the triangulation (Fig. \ref{fig:moving}(b)); then re-triangulate the hole left by the deleted tetrahedra (Fig. \ref{fig:moving}(c)). \emph{New point insertion}: insert the new point $p_{\text{new}}$ into the triangulation and add to the set $U$ all the rays stored in the conflicting tetrahedra (Fig. \ref{fig:moving}(d-f)). \emph{Rays removal}: for each tetrahedron of the entire triangulation remove the rays ending in $p_{\text{old}}$. \emph{Ray casting}: cast one ray for each ray in $U$.

In our case, whenever the 3D estimate of a point moves, we apply the Refinement Event Handler, before point addition and region growing, if and only if the point is inside the shrinked volume $D_{t_k}$ (see Section \ref{sec:manifold}), otherwise we do not move the point (this second case happens very rarely \cite{Litvinov_Lhuillier_13}).
\subsubsection{Complexity}
The number of rays involved in space carving algorithms is $O(F\cdot N^2)$ where $F$ and $N$ represent respectively the number of frames and the number of points in the triangulation \cite{Lovi_et_al_11}, and the number of tetrahedra in a 3D triangulation grows quadratically with the number of points ($O(N^2)$). 
In Table \ref{tab:ComStraight} we reported the complexities for each of the previous stage; since our implementation exploits the CGAL \cite{cgal} 3D triangulation data structure, the complexity of a single Ray casting, i.e., a cast of a single ray, is $O(N)$ in the general case, but we bound the size of the viewing ray, to avoid to include too far uncertain 3D points estimates, so the final complexity becomes $O(1)$  (see \cite[p.94]{yu2013automatic}).

From the analysis in the table is quite clear that this straightforward solution is not scalable, especially for the dependence between the number of rays and the number of processed frames. 

\begin{table}[t]
\caption{Complexity analysis; \emph{``-''} means not existing step.}
\label{tab:ComStraight}
\scriptsize
\centering
\begin{tabular}{lcccc}
\toprule 
Step                & straightforward     & K  & proposed     & window \\
                    & algorithm & heuristic & algorithm & heuristic \\
\midrule
Rays collection     &  $O(F\cdot N^2)$ & $O(N)$ &-&-\\
Weight collection    &-&- &  $O(N)$ & $O(N)$ \\
Vertex Removal      &  $O(N)$           & $O(N)$ &  $O(N)$           & $O(N)$ \\
New points insertion&  $O(F\cdot N^2 \cdot N)$ & $O(N)$ &  $O(F\cdot N^2 \cdot N)$ & $O(N)$ \\
Rays removal     &  $O(N^2\cdot F\cdot N^2)$ & $O(N^2)$ &-&-\\
Weight Update     &-&-&  $O(N)$ & $O(N)$ \\
Backward ray casting &-&-    &  $O(N^2\cdot F)$ & $O(N)$ \\
Ray casting     &  $O(N^2\cdot F)$ & $O(1)$ &  $O(N^2\cdot F)$ & $O(1)$ \\
\midrule
Overall complexity     &  $O(N^4\cdot F)$ & $O(N^2)$ &  $O(N^3\cdot F)$ & $O(N)$ \\
\end{tabular}
\end{table}

\subsubsection{Forgetting Heuristic}
Lovi et al. \cite{Lovi_et_al_11} proposed a \emph{forgetting} heuristic to limit the number of rays stored in each tetrahedron to a fixed number $K$, thus making the complexity independent from the number of the processed frames. 
However, we show in Section \ref{sec:experimental-results} that, when the points are moving,  the reconstruction is very inefficient even with this heuristic.

\subsection{The efficient approach}
\label{subsec:efficient_way}
Our contribution in this paper is an approach to deal with moving points, different from the straightforward variation of \cite{Lovi_et_al_11}. Indeed in our proposal, we avoid storing the list of rays inside each tetrahedron, and we just store the weight associated with it.
This allows the incremental reconstruction algorithm of Section \ref{subsec:incrementalManifold}, and, at the same time, we are able to bound the temporal complexity.

The main difficulty in the proposed approach is updating coherently the weights whenever a point moves, i.e., when the point is removed from the triangulation and added as a new point. As soon as the point is removed from the triangulation, we perform a backward ray casting with negative weights for each viewing camera such that the influence of the point is neglected. Then we remove the point, and we add a new vertex in the new position. Finally, we perform the ray casting from each viewing camera to the new point.

During both point removal and addition, we have to remove a set of connected tetrahedra from the triangulation and add a new one. 
Let $R = \{\Delta_1^R, \Delta_2^R, \dots, \Delta_{n_R}^R\}$ be the set of removed off tetrahedra and $A = \{\Delta_1^A, \Delta_2^A, \dots, \Delta_{n_A}^A\}$ the set of the new ones; their associated weights are respectively $W_R = \{w_1^R, w_2^R, \dots, w_{n_R}^R\}$ and $W_A = \{w_1^A, w_2^A, \dots, w_{n_A}^A\}$. The weights $W_R$ are known, while $W_A$ are those to be computed for the new tetrahedra, without recasting the visibility rays related to those tetrahedra.

Different approaches are possible: \emph{Mean value}: $w_i^A = \frac{1}{n_A}\sum_{k=1}^{n_R} w_{k}$; \emph{Weighted mean}: let $d_{i,j}$ be the Euclidean distances between the centroids of the $i$-th tetrahedron of $A$ and the $j$-th of $R$; then $w_i^A = \frac{\sum_{k=1}^{n_R}d_{i,k}^{-1}}{\sum_{k=1}^{n_R}d_{i,k}^{-1}}$; \emph{Minimum distance}: $w_i^A = w_{\bar{j}}^R$ such that $\bar{j} = \argmin_{j \in {1\dots n_R}}(d_{ij})$.

\begin{figure}[t]
\begin{center}
\begin{tabular}{cccc}
\centering
\includegraphics[width=0.21\columnwidth]{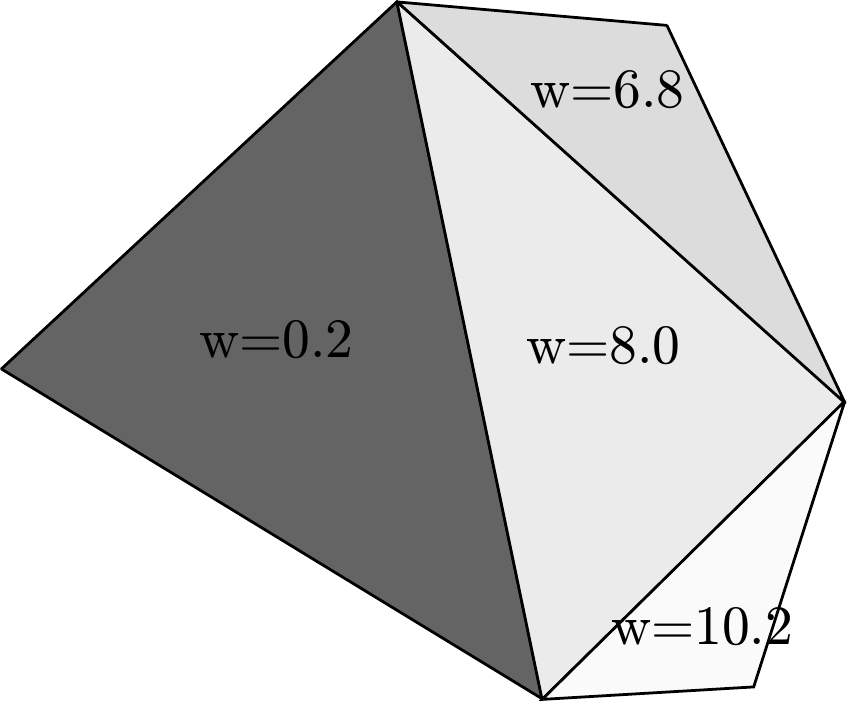}&
\includegraphics[width=0.21\columnwidth]{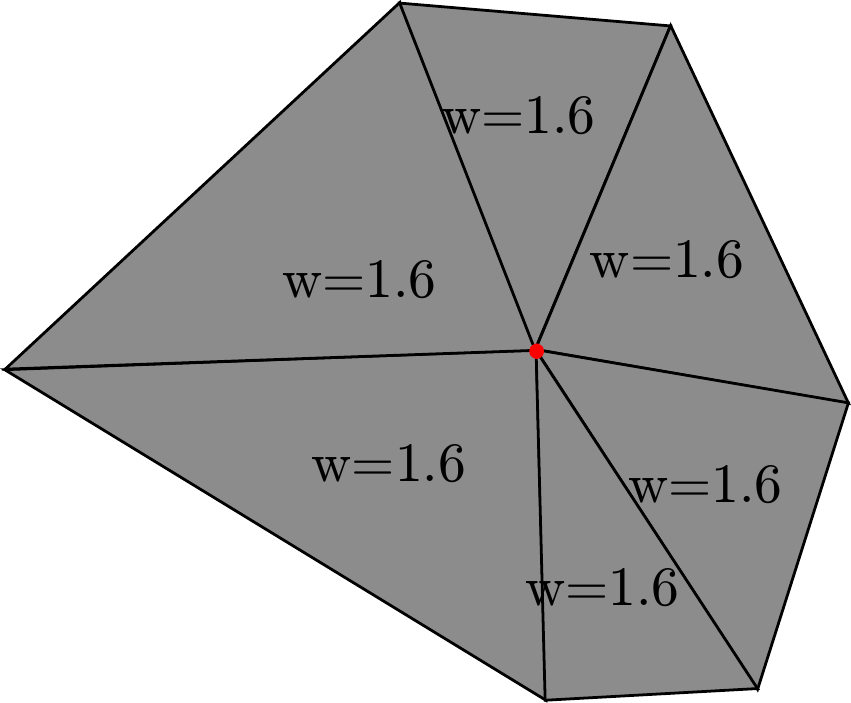}&
\includegraphics[width=0.21\columnwidth]{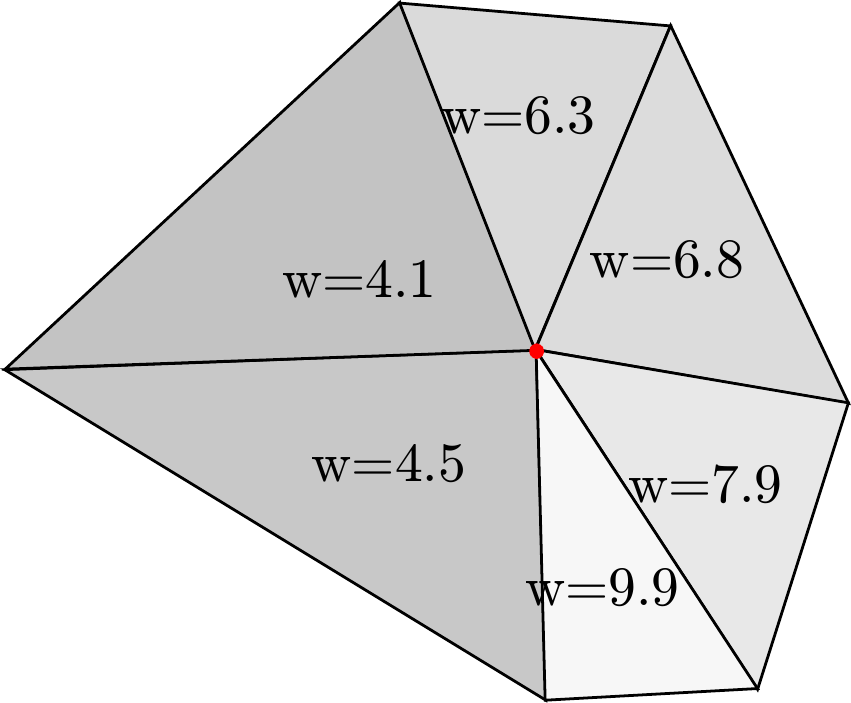}&
\includegraphics[width=0.21\columnwidth]{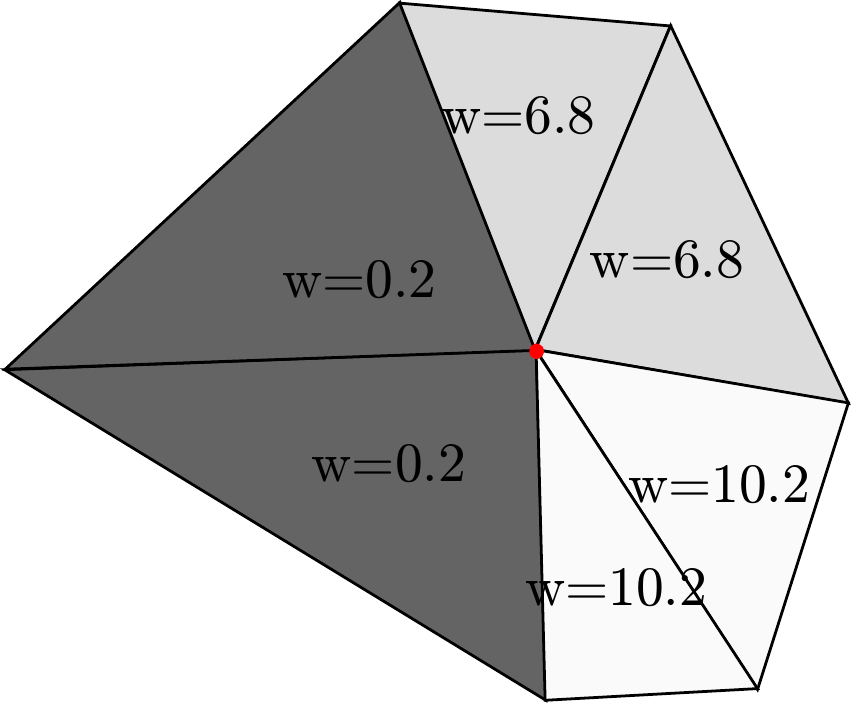}\\
(a) & (b) & (c) & (d)\\
\end{tabular}
\end{center}
\caption{2D example of moving point addition in the new position after point removal (a brighter region corresponds to a higher weight, i.e., higher probability to be carved).}
\label{fig:reconstrEx}
\end{figure}

Among these, the third solution gives a non-smooth outcome and, even if this seems counter-intuitive, it results to be more suitable for our purposes. The main reason is that it preserves the discontinuity between matter and free space. For instance in Fig. \ref{fig:reconstrEx}(a) we depict a 2D triangulation where we want to add a new point position; in Fig. \ref{fig:reconstrEx} (b), (c) and (d) we show the results of weights update after point addition with, respectively, Mean value, Weighted Mean value and Minimum distance approaches. 
It is clear that only Fig. \ref{fig:reconstrEx}(d) preserved the discontinuity, while in other cases becomes hard to distinguish between matter (lower weights) and free space (higher weights).

In case of very sparse data, the centroids of big tetrahedra, together with the associated visibility information, can be far from the newly added or moved points, and our update policy might lead to results far from the ideal solution, i.e., the straightforward approach discussed in Section \ref{subsec:straightforward_way}. Our algorithm overcomes this issue thanks to the use of the (so called) Steiner points added to the triangulation before the actual reconstruction is performed; this idea was already introduced in \cite{Litvinov_Lhuillier_13}.
We add Steiner points to the Delaunay triangulation every 5m along each axis so that they cover all the space that can be represented. The use of Steiner points limits the creation of very big tetrahedra, the visibility information becomes always local, and the update policy avoids drifts. Indeed, experimental results show good accuracy on varied scenes, even when lack of textures induces very sparse data.

\subsubsection{Complexity}
The complexity of the steps of our algorithm are reported in Table \ref{tab:ComStraight}. The main difference with respect to the straightforward algorithm is the replacement of the Rays removal to the weight update and backward casting which are the key of the gaining in computational complexity.
The proposed algorithm is thus $O(F\cdot N^2)$, so, in principle, the dependence with $F$ still remains and results in a non scalable solution. 

\subsubsection{Window Heuristic}
\label{subsub:window}
We are able to bound the complexity of our algorithm to $O(N^2)$ thanks to the following heuristic: instead of backward casting all the rays connecting the moving point to all the viewing cameras, we consider only the most recent cameras.
In this case the complexity of the ray casting becomes  $O(W\cdot N^2)$, where $W$ is the (constant) size  of the window (in our case $W = 15$), so the final complexity is $O(N^2)$.

\section{Experimental validation}
\label{sec:experimental-results}
To evaluate our approach, we tested the system on four different sequences of the KITTI dataset \cite{Geiger_et_al12} on a 4 Core i7-2630QM CPU at 2.2Ghz (6M Cache), with 6GB of DDR3 SDRAM.
The video stream was captured by a Point Grey Flea 2, which records $1392\text{x}512$ gray scale images at $10$ fps. The vehicle pose are estimated through a RTK-GPS and they are the initial input of our system together with the video stream. 

Among all the sequences we choose the 0095 (268 frames) and 0104 (313 frames) since they depict two different urban scenarios: the former shows a narrow environment where the building fa\c{c}ades are close to the camera, the latter captures a wide road.
We also tested our approach on sequences 03 (801 frames) and 04 (271 frames) from the odometry dataset: these videos provide a varied landscape mixing natural (trees and bushes) and man-made (houses, cars) features.


\begin{figure}[t]
\centering
\begin{tabular}{cc}
\includegraphics[width = 0.48\textwidth]{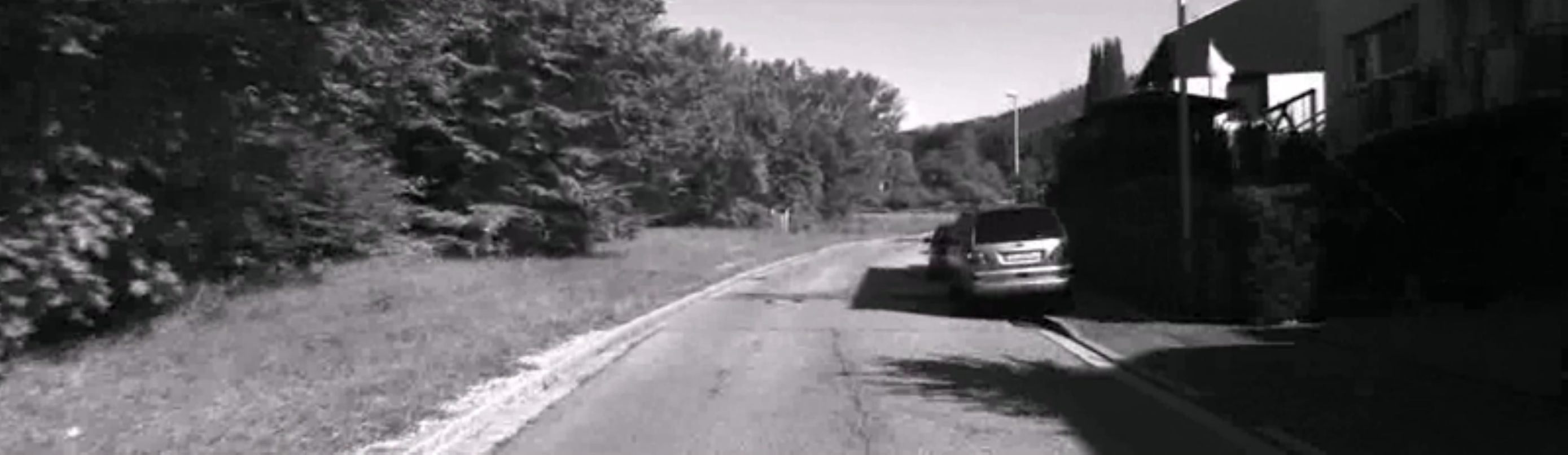}&
\includegraphics[width = 0.48\textwidth]{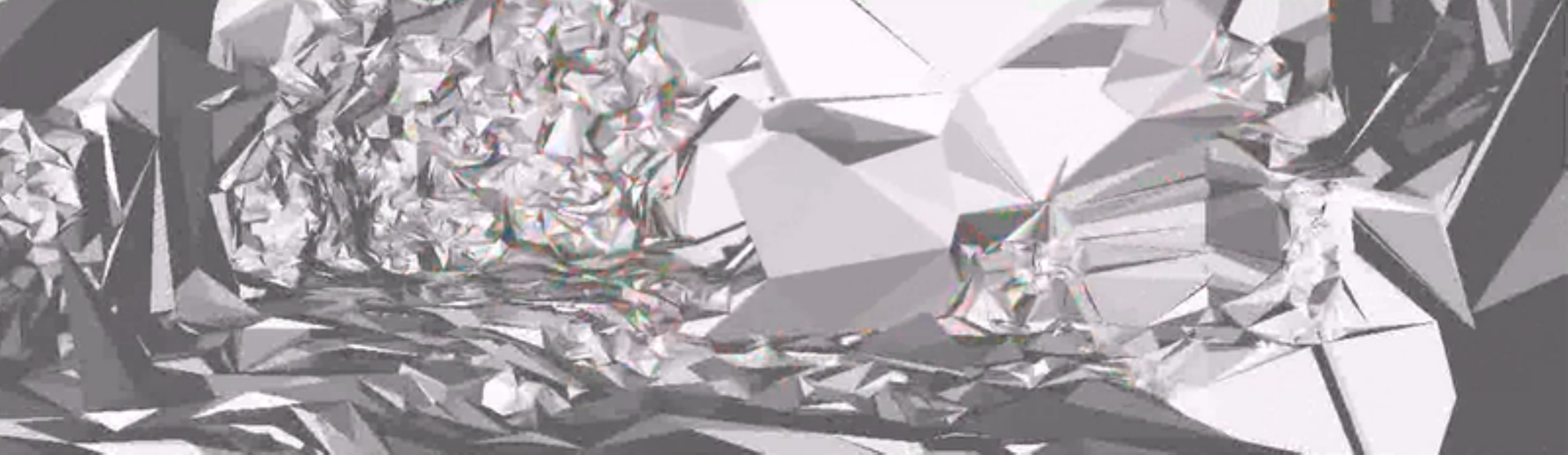}\\
\includegraphics[width = 0.48\textwidth]{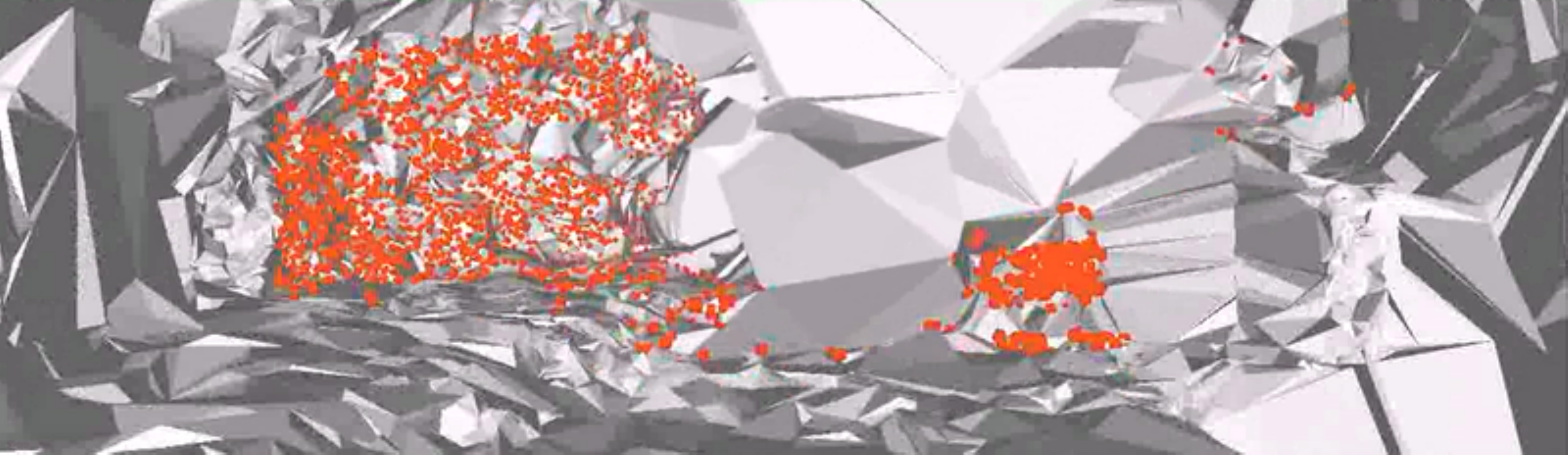}&
\includegraphics[width = 0.48\textwidth]{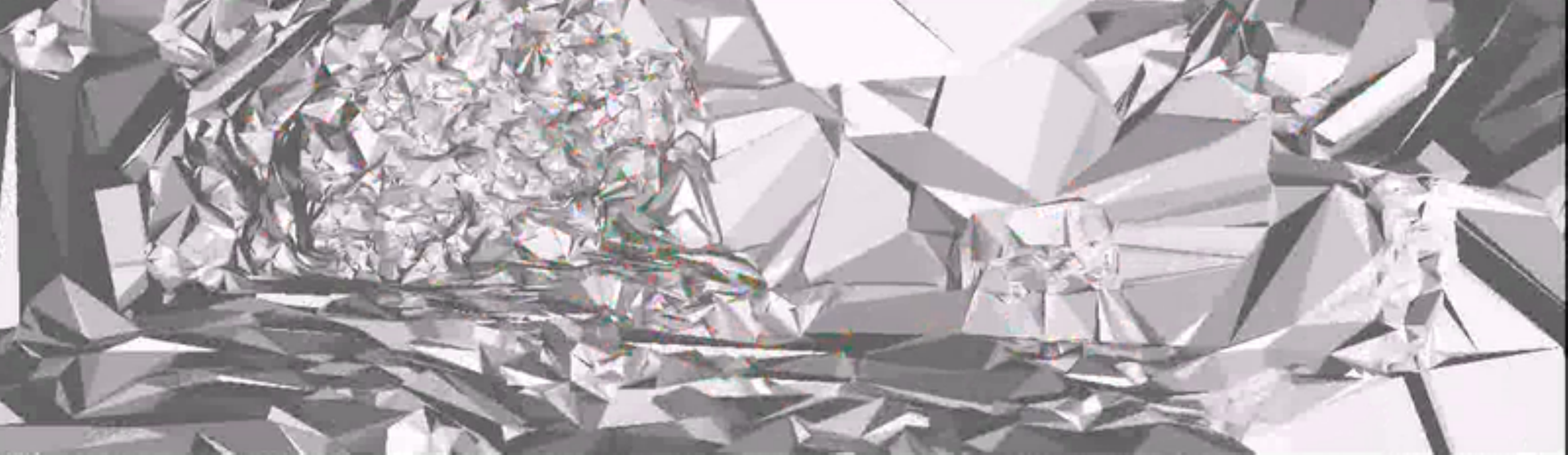}\\
\end{tabular}
\caption{Incremental reconstruction example. From up left to bottom right: original frame, before point positions update, points moved in the scene (red dots) and  manifold updated.}
\label{fig:exampleFr}
\end{figure}


To provide a quantitative evaluation we compared the reconstructed meshes with the very accurate point clouds measured by the Velodyne HDL-64E sensor in the KITTI dataset through the CloudCompare tool \cite{CloudCompare}.
This tool computes the reconstruction error as the average of the distances between each Velodyne point and the nearest triangle in the reconstructed mesh.

We evaluated the performance and the accuracy of the Lovi's approach against our three different updating policy.
As explained previously, no manifold incremental reconstruction approach deals with moving points, so a fair comparison results to be between the straightforward approach of Lovi, applied to manifold reconstruction (Section \ref{subsec:straightforward_way}) and our updating policies. In Fig. \ref{fig:exampleFr} we show an example of the reconstruction results before and after the red points has been moved in the Delaunay triangulation (see also the video at \url{http://youtu.be/_-q9sKjcOC0}).

\begin{figure}[t]
\centering
  \begin{tabular}{ccc}
    \centering
    \includegraphics[height=0.25\textwidth]{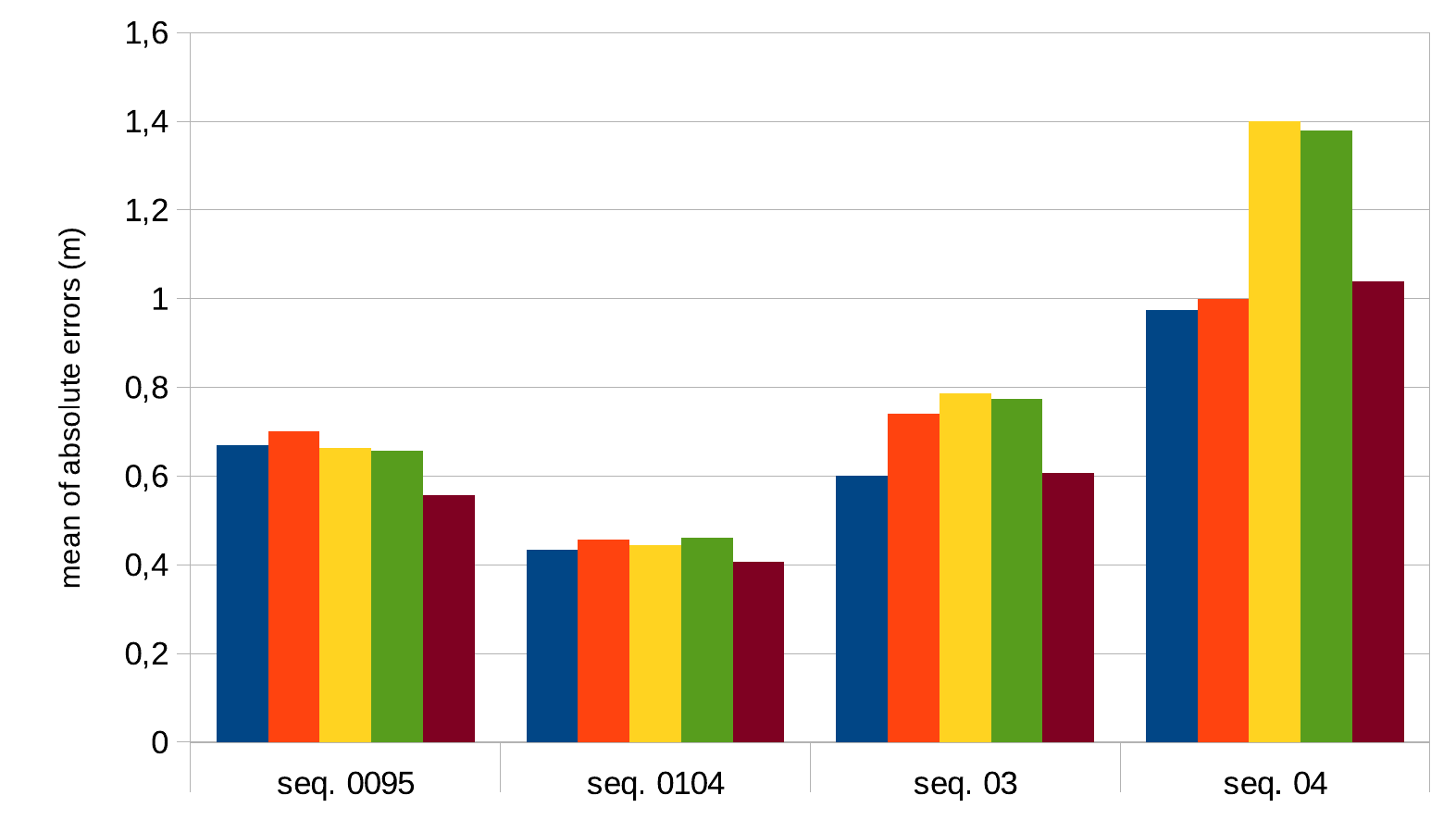}&
    \includegraphics[height=0.25\textwidth]{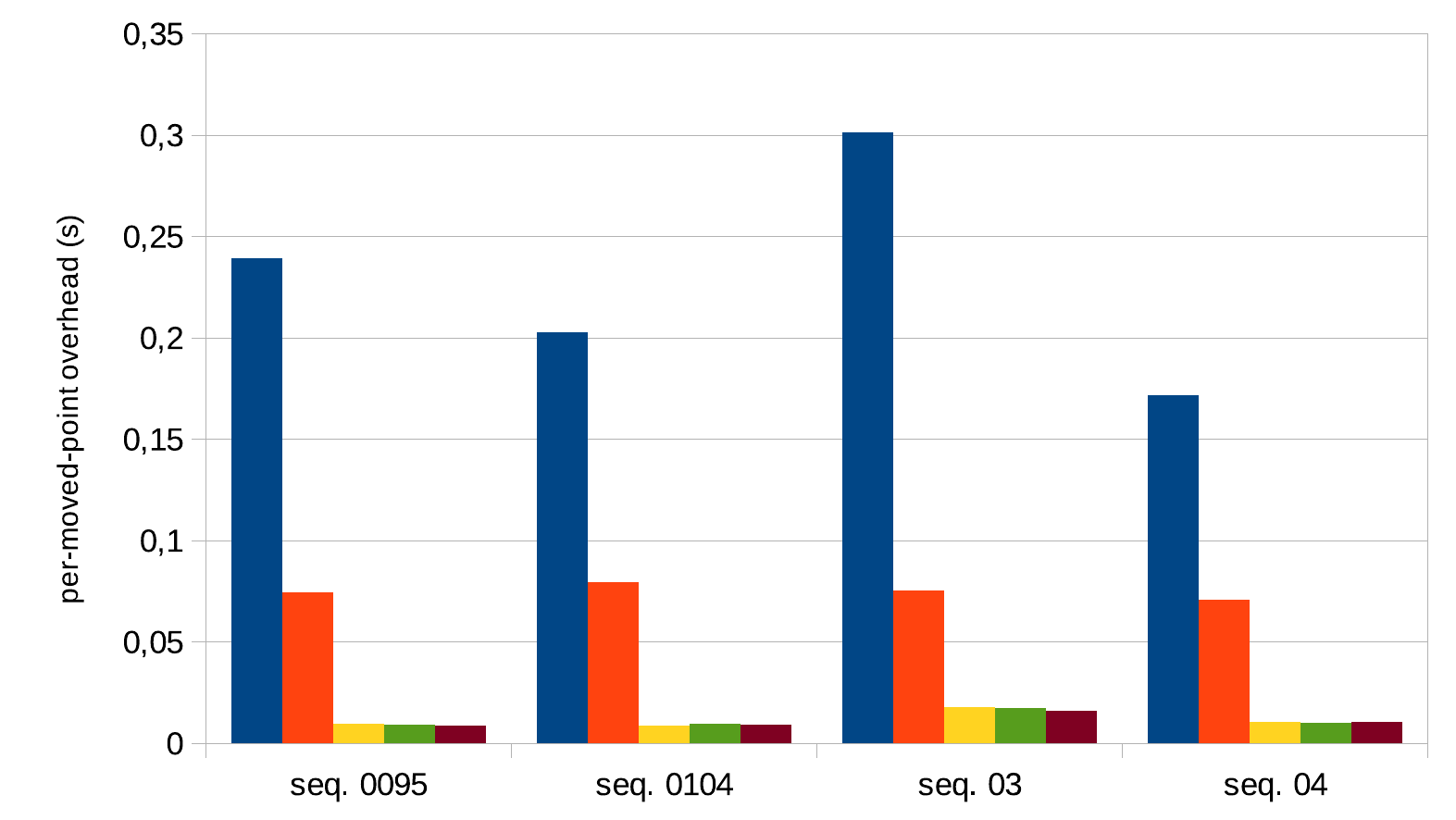}&
    \includegraphics[height=0.22\textwidth]{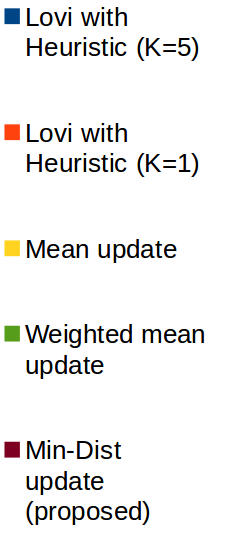}\\
    (a) Absolute errors (m).&
    (b) Per-point overhead (s).&\\
  \end{tabular}
  \caption{Experimental evaluation of the proposed approach with respect to Lovi's \cite{Lovi_et_al_11}.}
   \label{tab:results}
\end{figure}

Fig. \ref{tab:results} shows the results of the comparison where we applied the window heuristic (Section \ref{subsub:window}) to all the algorithms. In the case of Lovi's algorithm we applied the forgetting heuristic with $K=5$ and $K=1$, where $K$ is the number of viewing rays stored for each tetrahedron.
Fig.  \ref{tab:results}(a) shows that the accuracy of the proposed approach, i.e, moving point management through minimum distance weight updates, is comparable with respect to Lovi's proposal outcomes, where the algorithm with $K=5$ stores more information, so it performs better.
We compared our approach with respect to Lovi's method instead of the other incremental reconstruction algorithm presented in \cite{Litvinov_Lhuillier_13}; the reasons are twofold. 
In \cite{Litvinov_Lhuillier_13} Litvinov and Lhuiller does not deal with moving points, which is the main point addressed in this paper. Moreover, Litvinov and Lhuiller point out in \cite{litvinov_Lhiuller14} that the ideal solution for a manifold reconstruction algorithm is represented by the manifold including as much as free space tetrahedra as possible. Since the solution provided by Lovi et al. coincides with the (non-manifold) mesh containing all the free space tetrahedra, a reconstruction accuracy similar to Lovi's suggests that the reconstruction is near to the ideal solution. In some cases our algorithm reaches even better accuracy, this is due to the smoothing effect induced by our heuristic.

Fig. \ref{tab:results}(a) shows that the Minimum Distance always outperforms the other two updating schema as expected (see the Section \ref{subsec:efficient_way}).

In Fig. \ref{tab:results}(b) we report the time performance of the algorithms.
Let $T_{\text{mov}}$ and $T_{\text{non-mov}}$ be the overall processing time with and without moving points, and $N_{\text{mov}}$ be the number of the total points moves, e.g., if one point moves three times, $N_{\text{mov}}=3$.
The overhead introduced in the whole reconstruction process for each move of each point has been computed as $\frac{T_{\text{mov}} - T_{\text{non-mov}}}{N_{\text{mov}}}$.
The performance of the different update schema we presented in Section \ref{subsec:efficient_way} is very similar since the steps involved are basically the same: for each update on the Delaunay data structure, we iterate over the old tetrahedra to collect the weights, then we iterate over the new tetrahedra to set the new weights.
As expected by Section \ref{subsec:efficient_way}, our algorithm clearly outperforms Lovi's approach. Our updating schema is very efficient for two reasons. 
First, we only need to update locally the visibility, while Lovi's approach casts a ray for each visibility ray stored inside the tetrahedra.
Second, when we remove a point (first step of moving point management), we perform a ray casting backward to update only the convenient tetrahedra,  instead of iterating over the whole triangulation to remove the visibility rays involving the point moved as in \cite{Lovi_et_al_11}.

\section{Conclusion and future work}
\label{sec:conclusion}
In this paper we have shown that manifold reconstruction from sparse data with moving points is not a trivial task. To keep the Delaunay property valid when a point moves inside the Delaunay triangulation, we have to remove it and add a new point in the new position. This induces the removal of a set of tetrahedra, with the associated visibility information; then, we have to add a new set of tetrahedra with coherent visibility information; finally we have to update the visibility information in all the tetrahedra affected by the point move.

Existing solutions successfully applied for classic Space Carving, result to be inefficient and slow when applied in the manifold reconstruction setting. 
In this setting, we investigated different approaches to handle visibility information propagation, by updating the weight for each tetrahedron, which roughly represents  the number of ray intersections, and we proposed an efficient algorithm to conveniently update it. 
We tested our system with the KITTI dataset and it clearly outperforms the existing approach of Lovi et al. \cite{Lovi_et_al_11} for incremental manifold reconstruction.

Future works would include a photometric refinement of the manifold extracted incrementally, and an evaluation of the manifold quality on-the-fly, relying on the uncertainty information carried by the estimation of 3D points. 
A natural extension could also deal with the reconstruction of non-rigid shapes whose 3D points are moving.

\section*{Acknowledgements}
\footnotesize
This work has been partially funded by the SINOPIAE project, from the Italian Ministry of University and Research and Regione Lombardia, and the MEP: Maps for Easy Paths project funded by Politecnico di Milano under the POLISOCIAL program.
\bibliographystyle{splncs}
\bibliography{biblioEdgePoint}

\end{document}